\documentclass[9pt]{article}
\usepackage{spconf,amsmath,amssymb,graphicx}
\usepackage{multirow}
\usepackage{multicol}
\usepackage{booktabs}
\usepackage{graphicx}
\usepackage{textcomp}
\usepackage{xcolor}
\usepackage{color,soul}

\newcommand{\spacelength}{1em} 

\title {Grid-LOGAT: Grid Based Local and Global Area Transcription for Video Question Answering}
\name{%
\begin{tabular}{@{}c@{}}
Md Intisar Chowdhury$^{*}$ \qquad 
Kittinun Aukkapinyo$^{*}$ \qquad 
Hiroshi Fujimura$^{*}$ \\
Joo Ann Woo$^{}$ \qquad 
Wasu Wasusatein$^{}$ \quad 
Fadoua Ghourabi$^{}$ \quad
 \thanks{This paper is based on results obtained from a project, 
 JPNP23019, subsidized by the New Energy and Industrial Technology Development 
 Organization (NEDO), \hspace{5\spacelength} *Equal Contribution.}
\end{tabular}}

\address{$^{}$AWL, Inc.}

\begin{document}
\ninept
\maketitle
\begin{abstract}
In this paper, we propose a Grid-based Local and Global Area Transcription (Grid-LoGAT) system for Video Question Answering (VideoQA). The system operates in two phases. First, extracting text transcripts from video frames using a Vision-Language Model (VLM). Next, processing questions using these transcripts to generate answers through a Large Language Model (LLM). This design ensures image privacy by deploying the VLM on edge devices and the LLM in the cloud. To improve transcript quality, we propose grid-based visual prompting, which extracts intricate local details from each grid cell and integrates them with global information. Evaluation results show that Grid-LoGAT, using the open-source VLM (LLaVA-1.6-7B) and LLM (Llama-3.1-8B), outperforms state-of-the-art methods with similar baseline models on NExT-QA and STAR-QA datasets with an accuracy of 65.9\% and 50.11\% respectively. Additionally, our method surpasses the non-grid version by 24 points on localization-based questions we created using NExT-QA.
\end{abstract}

\section{Introduction}

In recent years, VideoQA task has gained significant attention for its potential applications in various real-world scenarios. For instance, by applying Vision Language Models (VLMs) to camera footage, it is possible to accurately comprehend the local environment and provide appropriate responses for tasks such as robotic operations~\cite{liu2024moka}, assistance in autonomous driving~\cite{park2024vlaad}, and analysis of physical store spaces~\cite{wang2024smart}. However, due to privacy and security concerns, it is often challenging to transmit video data (including latent representations) directly to the cloud for analysis using large-scale VLMs that require substantial computational resources. Conversely, on the edge site, the limitations of computational power make it difficult to handle the large models that are currently prevalent in cloud-based solutions. This paper proposes a technology that safeguards the privacy of image and video data by performing processing on the edge side, thereby eliminating the need to transmit such data directly to the cloud. Although there remains a potential risk that privacy-sensitive information may persist in the processed outputs, this study primarily concentrates on addressing this foundational step.

For VideoQA, Video-VLMs are one of the most generalized solution. These models compress entire video inputs into a predefined number of tokens, which are then processed alongside textual prompts to generate answers~\cite{wang2024qwen2,li2025llavaonevision}. While these models excel in VideoQA tasks, their inference processes are computationally expensive for edge sites, as they require processing large number of tokens for video. Multi-model VLM framework, as studied in ~\cite{li2023videochat, zhang2023simple}, is an appropriate approach that aligns with our use case. In this paradigm, pre-trained VLM generates captions for video frames, which an LLM uses along with the user's query to provide final answers. This approach is ideal for zero-shot and edge-cloud settings due to its modularity and flexibility.

\begin{figure}
    \centering
    \includegraphics[width=0.8\linewidth]{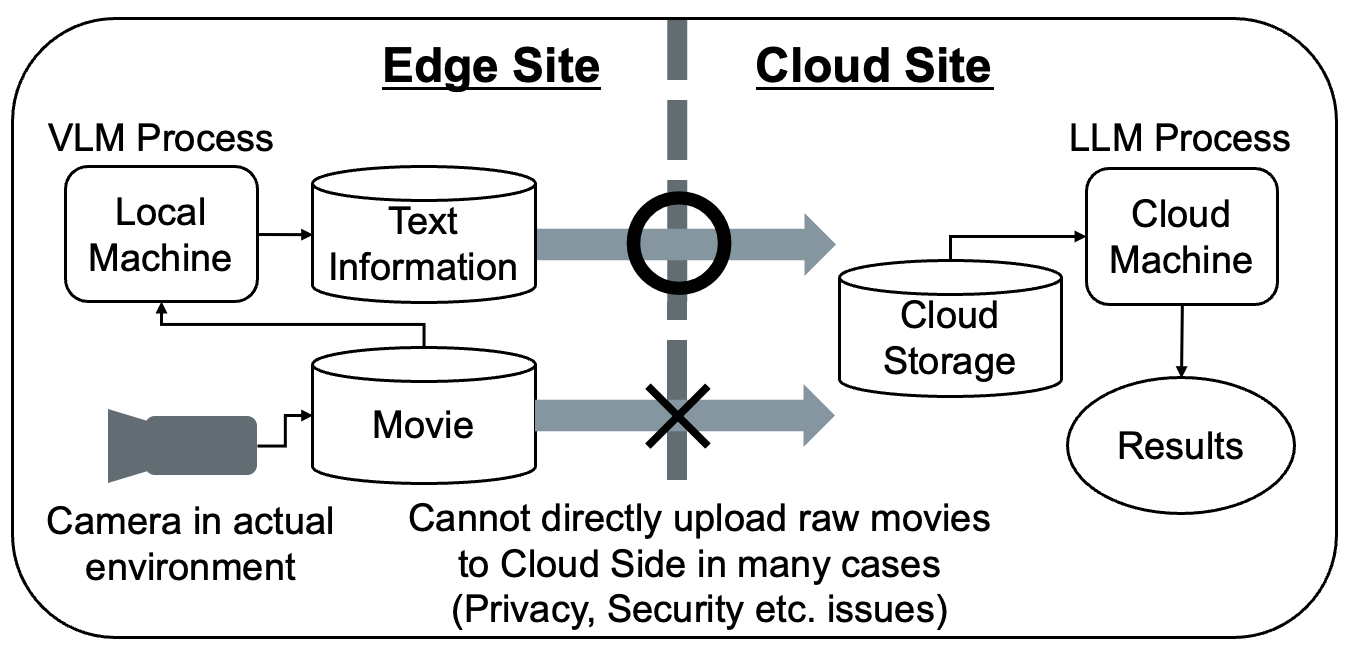}
    \caption{An instance of privacy-preserving video analysis with\\ Grid-LoGAT}
    \label{fig:sys_paradigm}
\end{figure}

One major limitation of the multi-stage VLM pipeline is its reliance on the performance of VLM, LLM, and the quality of the prompts used. Proprietary VLMs, such as GPT-4~\cite{achiam2023gpt} and Gemini 1.5 Pro~\cite{team2023gemini}, are highly capable of generating accurate captions, understanding scene context, reasoning, and grounding. However, they typically require uploading video and image data to cloud servers through APIs. On the other hand, open-source VLMs can preserve privacy and security through execution on the local site. However, these models often struggle with reliable scene analysis and grounding capabilities, failing to extract local details of the scene.


In this paper, we attempt to address these limitations of the VLM framework by proposing \textbf{Grid}-based \textbf{Lo}cal and \textbf{G}lobal \textbf{A}rea \textbf{T}ranscription for VideoQA, referred to as Grid-LoGAT. It extracts local transcription, which primarily captures detailed information from each grid cell in video frames, and global transcription, which captures the overall context of the video. By ensembling these transcriptions for each frame of the video, we obtain the final transcription. During the inference phase, the LLM solely relies on this local and global transcription. Grid-LoGAT is inherently built on lightweight, open-source VLM and LLM, making it suitable for processing visual data locally (e.g., on edge device) to ensure privacy. For answering user questions and queries, LLMs can be utilized on cloud servers, as text data is generally safer to handle remotely. In Fig.\ref{fig:sys_paradigm}, we illustrate the overall framework of the proposed VideoQA system, emphasizing privacy for images and videos.


\begin{figure*}
    \centering
    \includegraphics[scale=0.120]{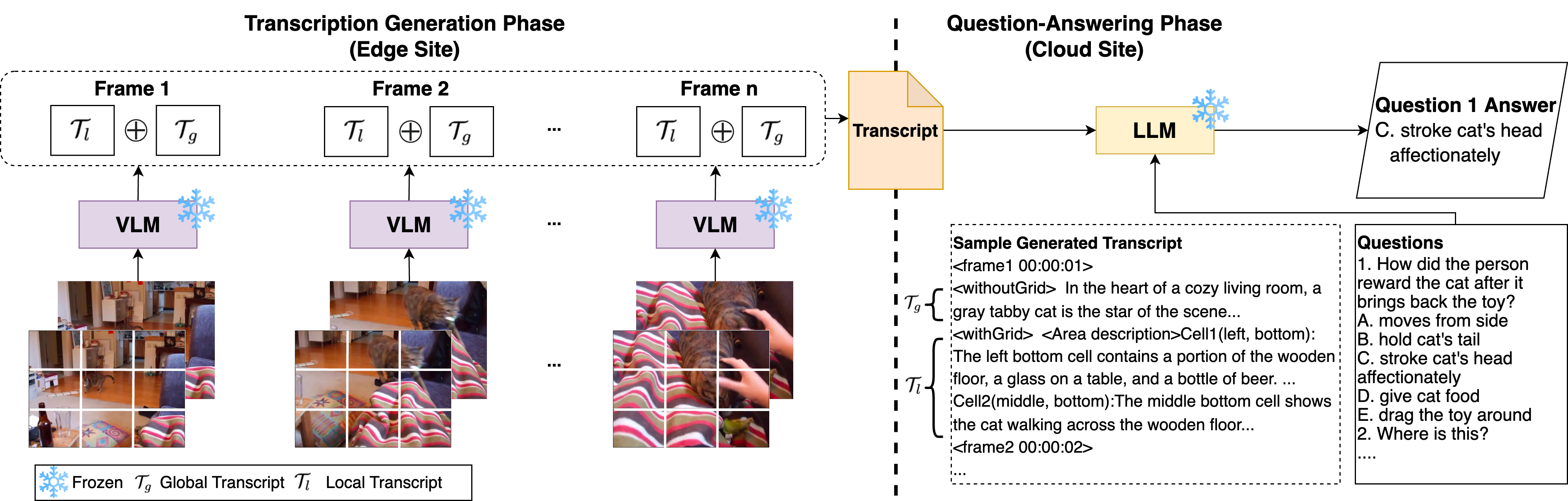}
    \caption{An overview of Grid-LoGAT. During the transcription generation phase, the VLM is fed with two variants of the image to capture local and global transcriptions. During the QA phase, the LLM refers only to the consolidated transcripts for free-form question answering. The motivation is that visual data can be handled securely on the edge site without being sent to the cloud, thus preserving privacy.}
    \label{fig:bigPicture}
\end{figure*}

\section{Literature Review}
\textbf{Video Question Answering} (VideoQA) through 
LLaVA~\cite{liu2024visual} is proposed to handle single-frame. Further research~\cite{li2025llavaonevision} extends the image VLM to support multi-frame and video data modalities.

While these methods excel at video analysis tasks like summarization and question answering, they often face limitations in extracting fine-grained information. For example, querying specific events at a particular frame or localizing certain individuals or objects in a specific frame can be challenging, as crucial detailed information may be compressed or lost during tokenization.

To address these challenges, recent approaches involve multi-stage video data processing using various off-the-shelf foundation models, eliminating the need for fine-tuning. For instance, the framework LLoVi, proposed in~\cite{zhang2023simple}, is designed for VideoQA and consists of a visual captioning model and an LLM. The visual captioning model generates textual descriptions of each frame, which are then fed to an LLM for summarization, refinement, and ultimately, question answering. VideoChat-Text~\cite{li2023videochat} utilizes a combination of video, image, and audio foundation models to process video streams into a human-readable format.

Beyond multi-stage processing, a single-stage zero-shot approach is proposed in~\cite{kim2024imagegridworthvideo}. In this method, several frames from a video are first concatenated into a grid layout. The image grid, together with a text prompt, is then fed into a pre-trained VLM for the VideoQA task.

Our work aligns with the aforementioned multi-stage, multi-model frameworks~\cite{li2023videochat, yu2024self}. Unlike monolithic approaches~\cite{wang2024qwen2, li2025llavaonevision} that rely on compressed visual tokens generated by video encoders, our approach first utilizes a lightweight auto-regressive VLM to extract fine-grained information from individual video frames. This fine-grained information subsequently enables more precise querying through an LLM across extended video sequences. Moreover, our proposed framework relies on open-source models and maintains a reasonable parameter count for the VLM, ensuring efficiency on edge devices.

\textbf{Visual prompting} enhances the visual grounding capabilities of VLMs, facilitating accurate object identification~\cite{yang2023set} and mitigating hallucinations~\cite{kuckreja2024geochat}. Visual prompting has been particularly explored in the image analysis domain. For example, in~\cite{yang2023set}, a segmentation model is used to partition each object in an image, followed by overlaying markers like numeric labels, alphanumeric tags, or bounding boxes. These annotated images, combined with a text query, are then input to GPT-4, resulting in enhanced grounding capabilities. Visual prompting based on spatial coordinates for querying specific regions of remote sensing images is explored in~\cite{kuckreja2024geochat, zhang2024earthmarker}. Research~\cite{nasiriany2024pivot} introduces an iterative visual prompting and optimization technique, where the VLM is queried repeatedly with visual prompts containing numerically annotated markers. The VLM selects the optimal option to complete the robotic task efficiently in an iterative manner. Building on the PIVOT framework~\cite{nasiriany2024pivot}, visual prompting for action localization is demonstrated in~\cite{wake2024open}. In this approach, multiple input images are concatenated and annotated with frame sequence numbers. These annotated visual prompt images are then input to GPT-4, which iteratively infers the action or event of interest, enabling effective localization.

While visual prompting has proven highly effective for improving grounding capabilities, particularly in image data, its application to local information extraction from videos remains largely under-explored. To address this gap, we propose a simple yet effective visual prompting strategy for local information extraction from video data. Our approach employs grid-based visual prompting to capture intricate local details. Unlike prior methods~\cite{li2023videochat, zhang2023simple, kim2024imagegridworthvideo}, which primarily focus on extracting only global information from each frame, our method advances by applying a specific grid pattern to individual video frames, enabling the extraction of detailed local information through pre-trained VLMs. The local and global information extracted by the VLM is then combined, enhancing accuracy in the question-answering task with the LLM based on these ensembled transcriptions.

\section{Method}
The system operates in two key phases. The first phase is the transcription generation phase. For any given video input, we utilize a VLM to extract both local and global transcriptions at the frame level. In this research, a transcription is defined as a structured video caption, which is composed of a timestamp and its corresponding dense caption describing each frame in detail. These extracted captions are then consolidated through element-wise concatenation (i.e., an ensemble of local and global information) to form a unified representation of the video content. The second phase is the question-answering phase. In this phase, we leverage an open-source LLM to answer any given question about the video by referencing the consolidated transcriptions of the first phase. An overview of the Grid-LoGAT is illustrated in Fig. \ref{fig:bigPicture}.

Our proposed system offers two key advantages. First, it enables privacy-preserved VideoQA. The system performs the VideoQA task solely based on the generated text transcript. Unlike existing works~\cite{kim2024imagegridworthvideo, yu2024self, xiao2022video}, our system does not rely on video data to generate answers. The generated transcript preserves the privacy of individuals who may appear in a video, as their visual appearance is not stored on the server site. Second, our system supports detailed information extraction through visual grid prompt. This approach improves the coverage of extracted information from each video frame, enriching the transcript for both general and location-oriented question answering. By leveraging textual transcripts, our system achieves privacy-preserved VideoQA while also improving accuracy.

\begin{table*}[]
\scriptsize
\centering
\caption{Evaluation result on NExT-QA and STAR-QA benchmarks}
\begin{tabular}{@{}l|l|c|c|cccc|ccccc@{}}
\toprule
\multicolumn{1}{c|}{\multirow{2}{*}{\textbf{Method}}} & \multicolumn{1}{c|}{\multirow{2}{*}{\textbf{Model}}} & \multirow{2}{*}{\textbf{\begin{tabular}[c]{@{}c@{}}LLM\\ Size\end{tabular}}} & \multirow{2}{*}{\textbf{\begin{tabular}[c]{@{}c@{}}Zero-Shot\\ Inference\end{tabular}}} & \multicolumn{4}{c|}{\textbf{NExT-QA}} & \multicolumn{5}{c}{\textbf{STAR-QA}} \\ \cmidrule(l){5-13} 
\multicolumn{1}{c|}{} & \multicolumn{1}{c|}{} &  &  & \textbf{Cas.} & \textbf{Tem.} & \textbf{Des.} & \textbf{\begin{tabular}[c]{@{}c@{}}Avg. \\ Acc\end{tabular}} & \textbf{Int.} & \textbf{Seq.} & \textbf{Pre.} & \textbf{Fea.} & \textbf{\begin{tabular}[c]{@{}c@{}}Avg. \\ Acc\end{tabular}} \\ \midrule
InternVideo~\cite{wang2022internvideo} & InternVideo & 1.3B & X & 48.0 & 43.4 & 65.1 & 59.1 & 43.8 & 43.2 & 42.3 & 37.4 & 41.6 \\
VideoChat2~\cite{li2024mvbench} & VideoChat2 & 7B & X & 61.9 & 57.4 & 69.9 & 61.7 & 58.4 & 60.9 & 55.3 & 53.1 & 59.0 \\ \midrule
SeViLA~\cite{yu2024self} & BLIP-2 & 2.85B & O & 61.3 & 61.5 & 75.6 & 63.6 & 48.3 & 45.0 & 44.4 & 40.8 & 44.6 \\
Vista-LLaMA~\cite{ma2023vista} & Vista-LLaMA & 7B & O & - & - & - & 60.7 & - & - & - & - & - \\
LLoVi~\cite{zhang2023simple} & GPT-3.5 & - & O & 69.5 & 61.0 & 75.6 & 67.7 & - & - & - & - & - \\
\multirow{4}{*}{IG-VLM~\cite{kim2024imagegridworthvideo}} & CogAgent~\cite{hong2024cogagent} & 7B & O & 52.3 & 47.3 & 65.9 & 52.8 & 39.8 & 47.4 & 40.5 & 43.6 & 44.4 \\
 & LLaVA 1.6 & 7B & O & 63.1 & 57.3 & 74.9 & 63.1 & 49.3 & 50.1 & 48.4 & 48.8 & 49.6 \\
 & LLaVA 1.6 & 13B & O & 61.6 & 55.7 & 70.8 & 61.2 & 51.5 & 52.0 & 51.0 & 51.8 & 51.7 \\
 & LLaVA 1.6 & 34B & O & 72.2 & 65.7 & 77.3 & 70.9 & 53.4 & 53.9 & 49.5 & 48.4 & 53.0 \\ \midrule
\textit{Ours} &  &  &  &  &  &  &  &  &  &  &  &  \\
w/o grid (global) & LLaVA 1.6 + Llama-3.1 & 8B & O & 65.8 & 57.4 & 75.8 & 64.9 & 45.9 & 51.8 & 52.1 & 44.5 & 48.6 \\
with grid (local) & LLaVA 1.6 + Llama-3.1 & 8B & O & 65.5 & 57.8 & 75.2 & 64.7 & 46.0 & 50.8 & 51.1 & 45.2 & 48.3 \\
local + global & LLaVA 1.6 + Llama-3.1 & 8B & O & 66.2 & 59.7 & 76.9 & 65.9 & 46.7 & 52.2 & 55.0 & 46.6 & 50.1 \\ \bottomrule
\end{tabular}
\label{table:benchmark}
\end{table*}

\begin{table*}[]
\scriptsize
\centering
\caption{Ablation studies on varying transcript configurations and grid layout with NExT-QA and STAR-QA dataset}
\begin{tabular}{@{}l|ccccccccc@{}}
\toprule
\multirow{2}{*}{\textbf{Scenario}} & \multicolumn{4}{c}{\textbf{NExT-QA}} & \multicolumn{5}{c}{\textbf{STAR QA}} \\ \cmidrule(l){2-10} 
 & \textbf{Cas.} & \textbf{Tem.} & \textbf{Des.} & \multicolumn{1}{c|}{\textbf{Avg. Acc}} & \textbf{Int.} & \textbf{Seq.} & \textbf{Pre.} & \textbf{Fea.} & \textbf{Avg. Acc} \\ \midrule
\textit{Transcript Configuration} &  &  &  & \multicolumn{1}{c|}{} &  &  &  &  &  \\
w/o grid (global) & 65.8 & 57.4 & 75.8 & \multicolumn{1}{c|}{64.9} & 45.9 & 51.8 & 52.1 & 44.5 & 48.6 \\
with grid (local) & 65.5 & 57.8 & 75.2 & \multicolumn{1}{c|}{64.7} & 46.0 & 50.8 & 51.1 & 45.2 & 48.3 \\
local + global & \textbf{66.2} & \textbf{59.7} & \textbf{76.9} & \multicolumn{1}{c|}{\textbf{65.9}} & \textbf{46.7} & \textbf{52.2} & \textbf{55.0} & \textbf{46.6} & \textbf{50.1} \\ \midrule
\textit{Grid layout (n \(\times\) m)} &  &  &  & \multicolumn{1}{c|}{} &  &  &  &  &  \\
2 \(\times\) 2 & 65.1 & 56.7 & 75.4 & \multicolumn{1}{c|}{64.2} & 43.7 & 50.2 & 50.6 & 42.1 & 46.6 \\
2 \(\times\) 3 & \textbf{65.5} & 57.8 & 75.2 & \multicolumn{1}{c|}{\textbf{64.7}} & \textbf{46.0} & \textbf{50.8} & 51.1 & \textbf{45.2} & \textbf{48.3} \\

3 \(\times\) 3 & 65.1 & \textbf{58.0} & \textbf{75.8} & \multicolumn{1}{c|}{64.6} & 44.6 & 49.4 & 51.1 & 44.7 & 47.6 \\ \bottomrule
\end{tabular}
\label{table:ablation}
\end{table*}

\begin{table*}[]
\scriptsize
\caption{Prompt templates: default system and user prompts for Grid-LoGAT. \(\mathcal{T}\), \(\mathcal{Q}\), and \(\mathcal{A}\) denote the generated video transcript, the input question or query, and the corresponding multiple-choice answers, respectively.}
\begin{tabular}{ll}
\hline
\textbf{Scenario}                                                     & \textbf{Prompt}                                                                                                                     \\ \hline
\begin{tabular}[c]{@{}l@{}}with  grid prompt \\ (system)\end{tabular} & \begin{tabular}[c]{@{}l@{}}You are an image analyzer tasked with providing dense  captions for  images overlaid with a 2x3 black color grid.\\ **Task:**\\ There are two tasks: 1. Overall description Ignore the grid lines in the image. Describe key details (objects, colors, \\ textures, etc.) for whole image. 2. Area description This  image's grid divides the image into  6 cells. And the 6\\ cells are located  (Cell1(left, lower), Cell2(middle, lower),  Cell3(right, lower), Cell4(left, upper),  Cell5(middle, upper),\\ Cell6(right, upper)). Describe key details (objects, colors,  textures, etc.) for each area centered on each cell.  Note \\ any patterns or relationships with  nearby cells.  \#\#Output Format: Overall description  Caption for whole \\ image. Area description Cell1(left, lower): Caption for Cell1, Cell2(middle, lower): Caption for Cell2 \\ Cell3(right, lower):  Caption for Cell3, Cell4(left, upper): Caption for Cell4, Cell5(middle, upper): Caption for Cell5\\ Cell6(right, upper): Caption for Cell6\end{tabular} \\
\begin{tabular}[c]{@{}l@{}}with grid prompt\\ (user)\end{tabular}     & Provide dense captions of this image.                                                                                             \\ \hline
\begin{tabular}[c]{@{}l@{}}w/o grid  prompt\\ (system)\end{tabular}   & LLaVA1.6 Default System Prompt                                                                                                          \\
\begin{tabular}[c]{@{}l@{}}w/o grid prompt\\ (user)\end{tabular}      & Carefully analyze this input image. Provide dense captions descripting in sentences.                                                                                                                                     \\ \hline
\begin{tabular}[c]{@{}l@{}}QA with LLM\\ (system)\end{tabular}        & \begin{tabular}[c]{@{}l@{}}**Objective**  Following is the frame-by-frame summarization results generated by a  vision-language \\ model. ** frame-by-frame summarization  start ** \(\{\mathcal{T}\}\) ** frame-by-frame  summarization ends **\end{tabular}                                                                                                                                                                                \\
\begin{tabular}[c]{@{}l@{}}QA with LLM\\ (user)\end{tabular}          & \begin{tabular}[c]{@{}l@{}}Answer the following question: Question: \(\{\mathcal{Q}\}\) YOU MUST CHOOSE the most  closest answers by  deductive\\ method  or  method of  elimination or from any factual information  provided in the transcription. Answer \\ options as follows:  Answer options: \(\{\mathcal{A}\}\) STRICTLY ADHERE TO  THE  FOLLOWING OUTPUT FORMAT. \\ Output  should be either  A or B or  C or D or E. Output format:  answer {[}Put answer here{]} answer\end{tabular}                                                                                                                                                         \\ \hline
\end{tabular}
\label{table:prompts}
\end{table*}

\subsection{Transcription generation phase}
For any video input \(V\), we first split it into \(N\) frames, represented as \(v = \{v_i\}_{i=1}^N\), 
where each \(v_i \in \mathbb{R}^{H \times W \times 3}\). Here, \(H\) and \(W\) are the height and width of the input frames respectively, and 3 depicts the RGB color channels. In this study, frames are extracted from the video at a rate of one frame per second (FPS). Consequently, the cardinality of \(v\), denoted as \(|v|\), equals the duration of the video in seconds.

While \(v\) represents the plain frames, we simultaneously construct a visual grid-based version of \(v\), denoted as \(v_{g}\), defined as:
\[
v_{g} = \{\mathcal{G}(v_i)\}_{i=1}^N,
\]
where the function \(\mathcal{G}\) overlays an \(n \times m\) grid marker on the input image. Here, \(n\) and \(m\) are the numbers of rows and columns, respectively.

Next, we construct two sets of text prompts, \(\mathcal{P}_{g}\) and \(\mathcal{P}_{l}\), for global and local information extraction, respectively. The global prompt \(\mathcal{P}_{g}\) is applied to the plain frames \(v\) to generate global captions for each frame. For all frames in \(v\), the global captions \(\mathcal{T}_{g}\) are obtained as follows:
\begin{equation}
    \mathcal{T}_{g} = \mathcal{F}(v, \mathcal{P}_{g}),
\end{equation}
where \(\mathcal{F}\) represents the pre-trained VLM that is prompted in a zero-shot fashion.

Simultaneously, the local prompt \(\mathcal{P}_{l}\) is applied to the grid-based frames \(v_{g}\) to generate captions for each grid cell of the grid images. The local captions \(\mathcal{T}_{l}\) for the grid-based images are obtained as:
\begin{equation}
    \mathcal{T}_{l} = \mathcal{F}(v_{g}, \mathcal{P}_{l}).
\end{equation}
The same text prompt templates \(\mathcal{P}_{g}\) and \(\mathcal{P}_{l}\) are leveraged for every input image \(v_i\) and \(\mathcal{G}(v_i)\), respectively.

For each frame, with local prompt the generated captions for all the \(n \times m\) grid cells are obtained as a single transcript. Thus, each of \(\mathcal{T}_{g}\) and \(\mathcal{T}_{l}\) consists of \(|v|\) elements, where each element corresponds to the caption generated for the \(i\)-th frame. The length of each caption may vary depending on the token output limit imposed by the model \(\mathcal{F}\). The concatenation of \(\mathcal{T}_{g}\) and \(\mathcal{T}_{l}\) is performed in an element-wise manner, ensuring that for each frame, the global and local captions are combined consistently. The element-wise concatenation operation of \(\mathcal{T}_{g}\) and \(\mathcal{T}_{l}\) is depicted as follows:
\begin{equation}
    \mathcal{T} = \mathcal{T}_{g} \oplus \mathcal{T}_{l},
\end{equation}
where \(\mathcal{T}\) is the text transcript of the input video data. To maintain the temporal consistency of events, we append timestamps and frame numeration to each element in the text transcript \(\mathcal{T}\).

\subsection{Question-Answering Phase}
In this phase, we leverage the LLM to answer questions about the input video by referring to the generated text transcripts \(\mathcal{T}\). Formally, for any input video \(V\), we define a question \(\mathcal{Q}\) and its corresponding multiple-choice answers \(\mathcal{A}\). Next, we construct a text prompt template \(\mathcal{P}\) through a function \(\mathcal{C}\) as follows:  
\[
\mathcal{P} = \mathcal{C}(\mathcal{T}, \mathcal{Q}, \mathcal{A}).
\]

Here, \(\mathcal{C}\) is a function that constructs the prompt template based on the contents of \(\mathcal{T}\), \(\mathcal{Q}\), and \(\mathcal{A}\).

To obtain the prediction, we prompt the LLM \(\mathcal{L}\) with the template \(\mathcal{P}\), as represented below:  
\[
A_{\text{pred}} = \mathcal{L}(\mathcal{P}).
\]  

In general, the output \(A_{\text{pred}}\) is the token prediction generated by \(\mathcal{L}\). Since the output can occasionally be noisy, we apply regular expression-based cleaning to extract the predicted answer.

\begin{table}[]
\centering
\caption{Evaluation on NExT-QA-Loc supplementary questions}
\begin{tabular}{c  c  c}
\toprule
\textbf{Scenario} &\textbf{Visual Prompt}  & \textbf{Accuracy (\(\%\))} \\ \midrule
w/o grid prompt    & No            & 53.3 \\
w/ grid prompt      & No            & 66.6 \\
w/ grid prompt      & Yes           & 77.5 \\ \bottomrule
\end{tabular}
\label{table:ablation_q}
\end{table}

\begin{figure*}
    \centering
    \includegraphics[scale=0.50]{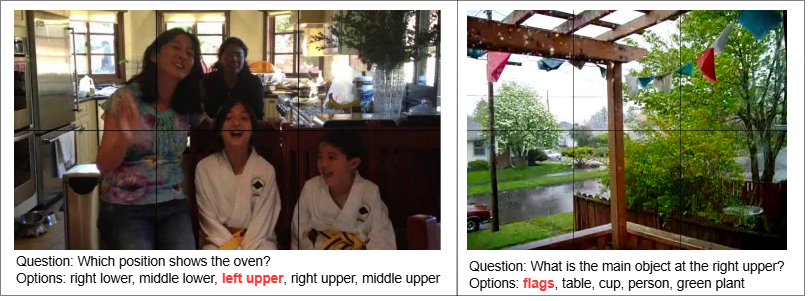}
    \caption{Instances of QA from NExT-QA-Loc. Ground-truth answers are depicted in red font.}
    
    \label{fig:gridqa}
\end{figure*}


\section{Experiments}


\subsection{Baseline models}
We leveraged the LLaVA-1.6-7B VLM~\cite{liu2024llava} to generate local and global transcriptions and the Llama-3.1-8B LLM~\cite{dubey2024llama} for the VideoQA task. Both models were chosen due to their open-source availability, reasonable parameter count, and fair comparison with methods using similar baseline models.  

For the VLM, we employed greedy sampling with a temperature of 1.0. The minimum and maximum numbers of generated tokens were set to 100 and 1000, respectively. During the question-answering phase, we set the sampling temperature for the LLM to 0.1 to ensure consistent and reproducible answers. The aforementioned settings were applied across all the datasets studied in this paper. Additionally, the prompt templates were kept identical across all datasets, as shown in Table \ref{table:prompts}. 

\subsection{Datasets}
We conducted experiments on two well-known VideoQA datasets, NExT-QA~\cite{xiao2021next} and STAR-QA~\cite{wu2024star}. Since all the experiments were conducted in zero-shot settings, we leveraged only the test set. The NExT-QA test set contains 1,000 videos and 8,563 multiple-choice QAs related to each video. The questions are mainly categorized into three types: causal, temporal, and descriptive. The STAR-QA test split consists of 955 videos with 7,377 multiple-choice QAs, classified into four types: interaction, sequence, prediction, and feasibility. Each multiple-choice QA (from both datasets) has only one correct answer, and we evaluated them using exact match accuracy on the official test split of both datasets.

Besides the official test splits, we prepared 120 localization-based questions from videos of the NExT-QA dataset, termed NExT-QA-Loc. These questions focus on the precise grounding and localization of people and objects. The purpose of this dataset is to fairly evaluate the effectiveness of grid-based prompting and the quality of local captions in general. We illustrated a few instances of these QAs for a clear understanding in Fig. \ref{fig:gridqa}.

\subsection{Result discussion}
\subsubsection{Comparison to other multi-model methods}
In Table \ref{table:benchmark}, for a fair and comprehensive comparison, we present the accuracies of relevant state-of-the-art methods in both zero-shot and video fine-tuned settings. All the methods shown in the table are composed of auto-regressive VLM models. Based on our empirical studies on grid-layout depicted in Table \ref{table:ablation} we choose \(2\times3\) as the candidate score for comparative studies.



\textbf{For NExT-QA,} our method achieved a considerable accuracy improvement (65.9\(\%\) vs. 63.1\(\%\)) in the zero-shot setting and with a similar baseline model (LLaVA-1.6-7B) as IG-VLM~\cite{kim2024imagegridworthvideo}. It should be noted that IG-VLM is a one-stage process, where during inference phase, the prompt, question, and grid of images are parsed simultaneously to the VLM. Our method was also comparable to multi-stage VLM system LLoVi~\cite{zhang2023simple} that leveraged much heavier proprietary models like GPT-3 (65.9\(\%\) vs. 67.7\(\%\)). Our method substantially outperformed SeViLA~\cite{yu2024self} (65.9\(\%\) vs. 63.6\(\%\)). It should be noted that, SeviLA relies on both filtered relevant video frames and text prompts as input, whereas our method relies solely on the text transcript. The accuracy of our method was also considerably competitive against several Video-VLMs~\cite{wang2022internvideo, li2024mvbench, ma2023vista} depicted in the Table \ref{table:benchmark}.

\textbf{For STAR-QA,} we maintained a competitive accuracy (50.11\(\%\) vs. 49.6\(\%\)) considering similar baseline model as \cite{kim2024imagegridworthvideo}. Our method considerably outperformed SeViLA~\cite{yu2024self} by 5.5\% (50.11\(\%\) vs. 44.6\(\%\)). However, since our methods used relatively smaller VLM and did not require fine-tuning on each dataset, it could not be fairly compared with other methods. For example, IG-VLM was also in a zero-shot setting but with a heavier baseline model (LLaVA-1.6-34B), and VideoChat2 \cite{li2024mvbench} which was adapted on video datasets.

In summary, in zero-shot settings, with baseline models of almost similar size, Grid-LoGAT outperformed state-of-the-art methods such as  LLoVi \cite{zhang2023simple}, IG-VLM \cite{kim2024imagegridworthvideo}, SeViLA \cite{yu2024self} and Vista-LLaMA \cite{ma2023vista} for the VideoQA task. 

\subsubsection{Ablation studies}

 \textbf{Ensemble of local and global transcripts} had a significant impact on accuracy across all the datasets studied. This improvement could be attributed to the ensembling effect of the VLM outputs and the inclusion of both dense local and global information. As shown in Table \ref{table:ablation}, a combination of local and global transcripts improved accuracy by approximately 1.5\(\%\) for the NExT-QA dataset, and 1.7\(\%\) for the STAR-QA dataset. This improvement came with the additional cost of executing the VLM twice, once with grid-based prompts and once without. However, as previously mentioned, the transcription generation step was executed only once.


\textbf{Efficacy of grid-based transcripts}
We have depicted the evaluation results on the NExT-QA-Loc in Table \ref{table:ablation_q}. First, we evaluated our proposed system without a rendered grid and grid-based prompts. It achieved an accuracy of 53.3\(\%\) as a baseline. Next, we configured the system with only grid-based prompts to enable the transcript to contain precise local information. It considerably resulted in an improved accuracy of \(66.6\%\). Finally, we combined both visual prompts and grid-based prompts, which resulted in the best accuracy of \(77.5\%\). It clearly shows the effectiveness of a visual grid.

\section{Conclusion}
In this research, we propose a multi-model, multi-stage VideoQA system coupled with grid-based visual prompts for the VideoQA task, which we term Grid-LoGAT. The system is inherently modular and built on open-source VLM and LLM. It first executes the VLM, prompted with a visual grid prompt, to extract both local and global information. This information is then ensembled to construct dense and rich video transcripts. Subsequently, an LLM leverages this information to answer questions by referring solely to the ensembled text transcription. Our proposed system offers several advantages due to its modular design. For instance, LLaVA-1.6-7B can be replaced with other VLMs depending on the computing budget at the deployment site. In the future, we aim to implement and benchmark such a system on edge platforms for privacy-preserving video content analysis.


\bibliographystyle{IEEEtran}
\bibliography{cite}

\end{document}